\documentclass[letterpaper]{article} 
\usepackage{aaai24}  
\usepackage{times}  
\usepackage{helvet}  
\usepackage{courier}  
\usepackage[hyphens]{url}  
\usepackage{graphicx} 
\usepackage{natbib}  
\usepackage{caption} 
\usepackage{algorithm}
\usepackage{algorithmic}

\usepackage{booktabs}
\usepackage{amsmath,amssymb,amsfonts}

\usepackage{newfloat}
\usepackage{listings}
\DeclareCaptionStyle{ruled}{labelfont=normalfont,labelsep=colon,strut=off} 
\floatstyle{ruled}
\newfloat{listing}{tb}{lst}{}
\floatname{listing}{Listing}
\graphicspath{{./Figures/}}

\title{Vision Transformer Off-the-Shelf: A Surprising Baseline for \\ Few-Shot Class-Agnostic Counting}
\author{
    Zhicheng Wang, 
    Liwen Xiao, 
    Zhiguo Cao, 
    Hao Lu
}
\affiliations{
    
    Key Laboratory of Image Processing and Intelligent Control, Ministry of Education; School of Artificial Intelligence and Automation, Huazhong University of Science and Technology, Wuhan 430074, China\\
    zhicheng\_wang@hust.edu.cn
}

\usepackage{bibentry}

\begin{document}

\maketitle

\begin{abstract}

Class-agnostic counting (CAC) aims to count objects of interest from a query image given few exemplars. This task is typically addressed by extracting the features of query image and exemplars respectively and then matching their feature similarity, leading to an extract-\textit{then}-match paradigm. In this work, we show that CAC can be simplified in an extract-\textit{and}-match manner, particularly using a vision transformer (ViT) where feature extraction and similarity matching are executed simultaneously within the self-attention. We reveal the rationale of such simplification from a decoupled view of the self-attention. The resulting model, termed CACViT, simplifies the CAC pipeline into a single pretrained plain ViT. Further, to compensate the loss of the scale and the order-of-magnitude information due to resizing and normalization in plain ViT, we present two effective strategies for scale and magnitude embedding. Extensive experiments on the FSC147 and the CARPK datasets show that CACViT significantly outperforms state-of-the-art CAC approaches in both effectiveness ($23.60\%$ error reduction) and generalization, which suggests CACViT provides a concise and strong baseline for CAC. Code will be available.

\end{abstract}
\section{Introduction}
Object counting aims to estimate the number of objects from a query image. Most prior object counting approaches target a specific domain, \textit{e.g.}, crowd~\cite{zhang2015cross,shu2022crowd,zou2021coarse}, plant~\cite{lu2017tasselnet,madec2019ear}, and car~\cite{onoro2016towards}. They often require numerous class-specific training data to learn a good model~\cite{wang2020nwpu}. In contrast, Class-Agnostic Counting (CAC), whose goal is to estimate the counting value of arbitrary categories given only few exemplars, has recently received much attention due to its potential to generalize to unseen scenes and reduced reliance on class-specific training data~\cite{lu2019class,famnet,bmnet,countr}. 

\begin{figure}[t]
\centering
\includegraphics[width=1\columnwidth]{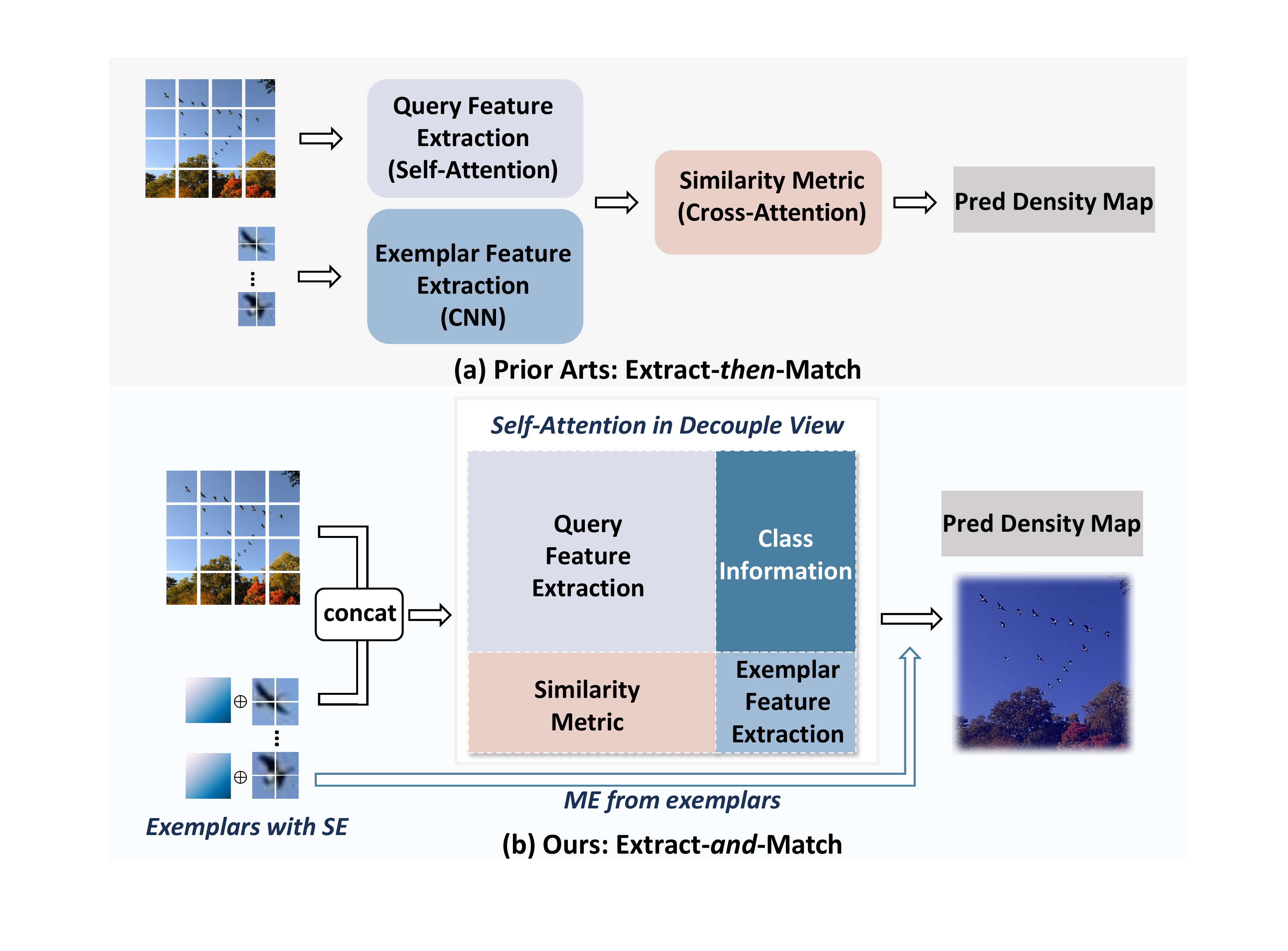} 
\caption{High-level ideas between prior arts and ours. (a) Previous ViT-based class-agnostic counting framework follows the extract-\textit{then}-match paradigm with unshared feature extractors (e.g., a ViT and a CNN) for the query image and the exemplars and post-matching such as cross-attention after feature extraction; (b) Our ViT-based framework follows an extract-\textit{and}-match paradigm using self-attention in a decoupled view, with additional aspect-ratio-aware scale embedding (SE) and the order-of-magnitude embedding (ME) for compensating the information loss of the scale in ViT.}
\label{fig1}
\end{figure}

CAC is first introduced by Lu \textit{et al.}~\cite{lu2019class}, which is by default formulated as a template matching problem, leading to an extract-\textit{then}-match paradigm. Previous models~\cite{famnet,bmnet,spdcn} use shared CNN for query images and exemplars feature extraction, as the bottom-up feature extraction approach of the CNN can adapt to images of entirely different sizes. Witnessing the ability of marking the responses on the attention map by cross-attention mechanism, some models such as CounTR~\cite{countr} employs cross-attention to match the features of query image and exemplars. However, in CounTR the query feature and exemplar feature are embedded separately by a ViT and a CNN, and the matching part is achieved by an extra cross-attention stage. This strategy introduces much redundancy and task-specific designs, which is not in line with the trend of task-agnostic foundation models.

Recently, the computer vision community has witnessed great success with plain ViT in large multi-modal architectures~\cite{touvron2023llama, yu2022metaformer}. Soon much work emerges for better adaptation of ViT on downstream vision tasks, such as object detection~\cite{vitdet, detrvit}, pose estimation~\cite{vitpose, xu2023vitpose} and image matting~\cite{ViTMatte}. As a template matching task, CAC is essentially suitable for using ViT with its attention mechanism; however, there is little focus on the adaptation of ViT on CAC task. 

In this work, we share insights that the attention mechanism in plain ViT has the ability to extract the features for both the query image and the exemplars and perform feature matching for them. By grouping the query and exemplar tokens into concatenation and feeding them to a plain ViT, the self-attention process in ViT can be divide into two groups of self-attention, and two groups of cross-attention. The former self-attentions are to extract features for the query image and the exemplars, while the latter cross-attentions contains the matching process between the query image and the exemplars. Therefore, without multiple feature extractors or extra post-matching, it produces a novel extract-and-match paradigm. Compared with prior arts, the extra attention from the query feature to the exemplars would further provide additional class information to the query image in this paradigm, enabling better perception of objects.Based on this idea, we propose a framework for CAC that mainly contains a single pretrained ViT, which verifies the feasibility of plain ViT for CAC task.

For better adaptation of ViT to the specific CAC task, we introduce more insights closely related to CAC task in our model design. Specifically, we observe that certain restrictions or functions such as resizing and $\tt softmax$ normalization within this architecture can result in the loss of scale information and the order of magnitude of counting values. First, the exemplars must be resized to fit the ViT input, which introduces size ambiguity during matching. Prior CNN-based models~\cite{bmnet} attempt to compensate for the scale information with scale embedding for exemplars; however, they neglect the information of aspect ratios, which is crucial for classes with abnormal ratios. This is largely overlooked in the existing literature. Second, the attention map with $\tt softmax$ function can represent the relative distribution of objects in the query image and therefore weakens the awareness of the model to the number of objects. We address this by restoring the magnitude order in the normalized attention map. Both the proposed scale embedding and magnitude embedding are easy to implement. By infusing the scale and the magnitude information into the plain ViT architecture, we acquire a surprisingly simple yet highly effective ViT baseline for CAC. The resulting model, termed CACViT, fully leverages the self-attention mechanism in ViT while also being tuned to mitigate the defects of this architecture in this task.

Experiments on the public benchmark FSC147~\cite{famnet} show that CACVit outperforms the previous best approaches by large margins, with relative error reductions of $19.04\%$ and $23.60\%$ on the validation and test sets, respectively, in terms of mean absolute error. Its cross-dataset generalization is also demonstrated on a car counting dataset CARPK~\cite{carpk}. We also provide extensive ablation studies to justify our propositions.

In a nutshell, our contributions are three-fold:
\begin{itemize}
\item A novel extract-and-match paradigm: we show that simultaneous feature extraction and matching can be made possible in CAC;
\item CACViT: a simple and strong ViT-based baseline, sets the new state-of-the-art on the FSC-147 benchmark;  
\item We introduce two effective strategies to embed scale, aspect ratio, and order of magnitude information tailored to CACViT. 
\end{itemize}

\section{Related Work}
The task of CAC is composed of two main components: feature extraction and feature matching. We first review each component in previous counting models, then discuss jointly feature extraction and matching in the other fields. 

\textbf{Feature Extraction in Class-Agnostic Counting.} 
The investigation of feature extraction in counting first began with class-specific counting~\cite{abousamra2021localization,cao2018scale,he2021error,idrees2018composition,laradji2018blobs,cheng2022rethinking}. In class-specific counting, most works are designed to address the challenges posed by quantity variance and scale variance. For class-agnostic counting, the core of feature extraction include unified matching space apart from challenges as above. To obtain a unified matching space, most previous work~\cite{famnet,bmnet,safecount} uses the shared CNN-based feature extractors for query images and exemplars. CounTR~\cite{countr}, which first introduces the ViT for feature extraction in CAC, uses different feature extractors for the query images (a ViT) and exemplars (a CNN). Hence, a two-stage training scheme is used for unifying the feature space. 

\begin{figure*}[t]
\centering
\includegraphics[width=\linewidth]{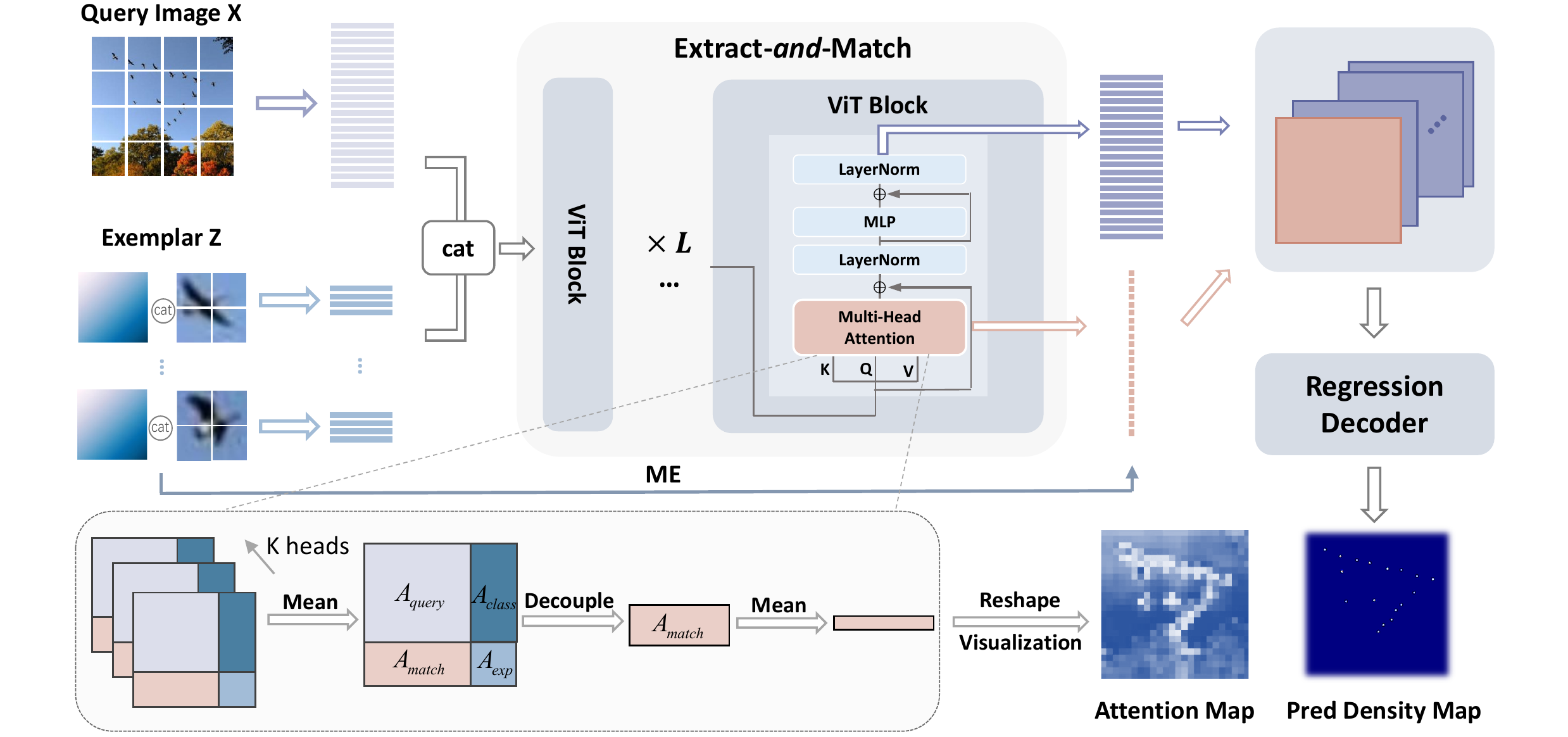} 
\caption{The framework of CAC Vision Transformer (CACViT). A query image and exemplars with scale embedding are spilt into patches to form tokens. Then the flattened tokens are concatenated and 
fed into the transformer encoder. Afterward, the output feature of query image and similarity metric from attention map with magnitude embedding (ME) are concatenated for regression. Finally, a regression decoder predicts the density map. It noted that the attention map is similar to density map.
}
\label{fig:pipeline}
\end{figure*}

\textbf{Feature Matching in Class-Agnostic Counting.} Compared with feature extraction, matching strategies in CAC have garnered more attention. The key points of the matching include the following two: 1) robustness to appearance variance, and 2) ability to characterize quantity levels. In the early attempt, naive inner product~\cite{famnet,yang2021class} is used, which is not robust to the appearance variance of objects to be counted. Shi \textit{et al.} ~\cite{bmnet} developed a bilinear matching network (BMNet) that expands the fixed inner product to a learnable bilinear similarity metric, which improves the robustness compared with the inner product. The recent ViT-based model CounTR~\cite{countr} uses cross-attention for matching, which seems a natural choice for a transformer-based solution at first glance. However, we show that, in our plain ViT model CACViT, we can perform feature matching at the same time of extracting features by self-attention.

\textbf{Jointly Feature Extraction and Matching.} 
For template matching and multi-modal tasks, feature extraction and matching are two main components. In tracking and detection tasks, MixFormer network~\cite{mixformer} and FCT network~\cite{crossdet} were proposed to enhance the correlation between the target object and the image, thereby obtaining enhanced features for localization head. In multi-modal tasks, ViLT~\cite{ViLT} strengthens the interaction between text and image during the feature extraction stage, resulting in efficient multi-modal features that benefit the performance of downstream tasks. To the best of our knowledge, we are the first to simultaneously consider feature extraction and matching in CAC, and we provide a decoupled analysis of the feasibility of this paradigm in the CAC, thereby streamlining the workflow of CAC task.

\section{Class-Agnostic Counting Vision Transformer}

In this section, we first introduce the overview of our framework for CAC, named Class-Agnostic Counting Vision Transformer (CACViT). Then, we provide an analysis of why the pure self-attention layers could serve as the substitution for previous architectures. After the ViT-style model is established, we spot the problem of scale information loss when directly applying ViT, and provide customized solutions for scale compensation.

\subsection{Overview of Approach}

The pipeline of CACViT is presented in Figure~\ref{fig:pipeline}. The query image $\mathcal{X}$ and corresponding exemplars $\mathcal{Z}$, which are resized to a fixed exemplar size, are split into tokens. Then they are concatenated and processed by self-attention layers. Afterward, the output feature of the query image and the similarity map from the last attention map are concatenated to form the final density map. The $\ell _2$ loss is adopted to supervise the density map. 

\subsection{A Decoupled View of Self-Attention}

We follow the standard formulation of self-attention in CACViT. Yet, in the presence of both query and exemplar tokens, we have a different interpretation of self-attention in CAC.

We first revisit self-attention in transformer. Given an input token sequence $\mathcal{I}$, we normalize it and transform it to a triplet of $\mathbf{Q}$, $\mathbf{K}$ and $\mathbf{V}$, through linear layers. Then we employ the scaled dot-product attention mechanism to compute the attention values between the queries $\mathbf{Q}$ and keys $\mathbf{K}$. Each output token is a weighted sum of all tokens using the attention values as weights, formulated as:
\begin{equation}
\tt Attention (\mathbf{Q},\mathbf{K},\mathbf{V})=\tt softmax (\mathbf{Q}^\top \mathbf{K}/\sqrt{D})\mathbf{V}\,,
\end{equation}
where the attention map $A =\tt softmax (\mathbf{Q}^\top \mathbf{K}/\sqrt{D})$.

In the presence of both query and exemplar tokens, we have $A \in \mathbb{R}^{ (M +  M_z)\times (M + M_z)}$, where $M$ and $M_Z$ are the token number of the query image and the token number of exemplars, respectively. Note that, with the help of skip connection~\cite{he2016deep} in the self-attention block~\cite{vaswani2017attention}, the physical meanings of the query tokens and exemplar tokens are preserved through layers, which is a prerequisite of our approach and analysis. 

By considering $A$ directly, it is not intuitive to see how the self-attention mechanism can function as the feature extractor and matcher. Here we provide a fresh look at self-attention in CAC with a decoupled view. 

First, we can split the token-to-token attention map $A$ into $4$ sub-attention maps as follow:
\begin{equation}
A = \begin{bmatrix}
  A_{query} &A_{class} \\
  A_{match}&A_{exp}
\end{bmatrix}\,,
 \end{equation}
where $A_{query} \in \mathbb{R}^{M \times M}$, $A_{class} \in \mathbb{R}^{M \times M_Z}$, $A_{match} \in \mathbb{R}^{M_Z \times M}$, and $A_{exp} \in \mathbb{R}^{M_Z \times M_Z}$. 

\begin{figure}[t]
  \centering
  \includegraphics[width=\linewidth]{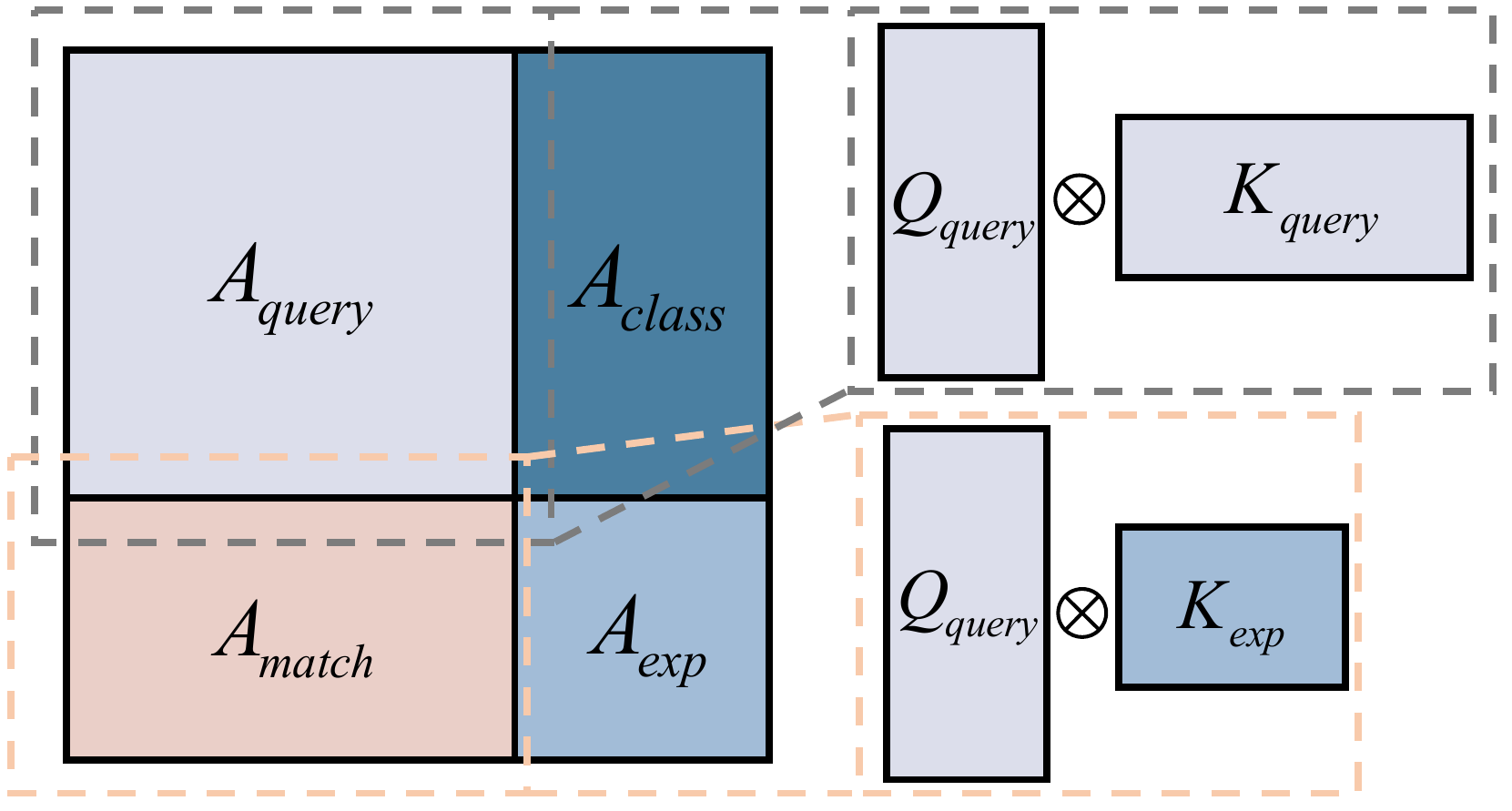}
  \caption{Decoupled view of self-attention in CACViT. The top-left $A_{query}$ can be regarded as self-attention of the query image. The bottom-left $A_{match}$ can be interpreted as cross-attention between query images and exemplars, despite being implemented in self-attention. }
  \label{fig:self}
\end{figure}

As Figure ~\ref{fig:self} shows, $A_{query}$ can be regarded as the self-attention map of the query image, which represents the most significant information within the image relevant to itself. $A_{query}$ performs feature extraction for the query image identical to the ViT or the CNN backbone in previous work. Similarly, $A_{exp}$ represents the self-attention map of the exemplars, which works as the feature extraction for the exemplars as in previous work.

$A_{match}$ shown in Figure~\ref{fig:self} can be regarded as cross-attention between the query image and exemplars, which thus can replace cross-attention layers used in previous ViT-based methods, taking the query image as the query and the exemplars as the key. In this way, $A_{match}$ can perform the feature matching. In summary, compared with previous work, self-attention in CACViT executes feature extraction and matching in a simple and natural way. 

In addition to the matching and extraction, we acquire an additional $A_{class}$. It is noted that the meaning of $A_{class}$ is different from $A_{match}$. Every row $i$ in $A_{class}$ represents the attention from query token $i$ to exemplar tokens. In the shallow layers, due to weak self-attention ability, query tokens with foreground objects can more quickly acquire corresponding category information through additional exemplar tokens. Conversely, background receives less attention from exemplar tokens. Thus, in the shallow layers, $A^T_{class}$ is similar to $A_{match}$, which focuses on the foreground objects. As the layers get deeper, the self-attention ability gets better. For query tokens containing foreground information, due to the repeated objects in the query image, self-attention will be stronger compared to query tokens containing background information. Moreover, since the total attention of the query image to itself and to the exemplar remains fixed, query tokens with background information will exhibit stronger attention to exemplars than query tokens with foreground information. As a result, the transposed $A_{class}$ will focus on the background region. 

\begin{figure}[t]
  \includegraphics[width=\linewidth]{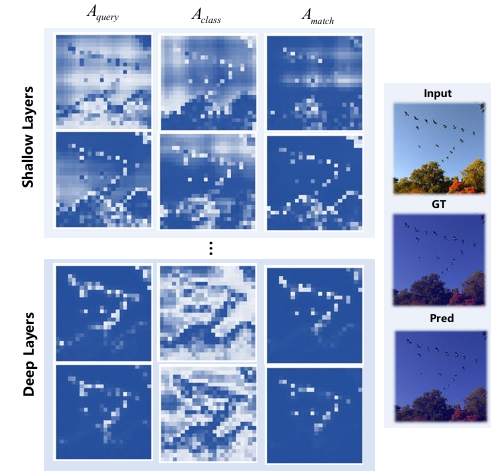}
  \caption{Visualizations of attention maps in a decoupled view for different layers. (a) $A_{query}$ and $A_{match}$ highlight the foreground and suppress the background. (b) In the shallow layers, $A_{class}$ favors foreground; but in the deep layers, $A_{class}$ highlights background. 
  }
  \label{fig:layerattention}
\end{figure}

The visualization results of decoupled attention are shown in ~\ref{fig:layerattention}, which verify our analysis. The more detailed analysis is on the supplementary.

\subsection{Scale and Magnitude Priors}
\begin{figure}[t]
  \includegraphics[width=\linewidth]{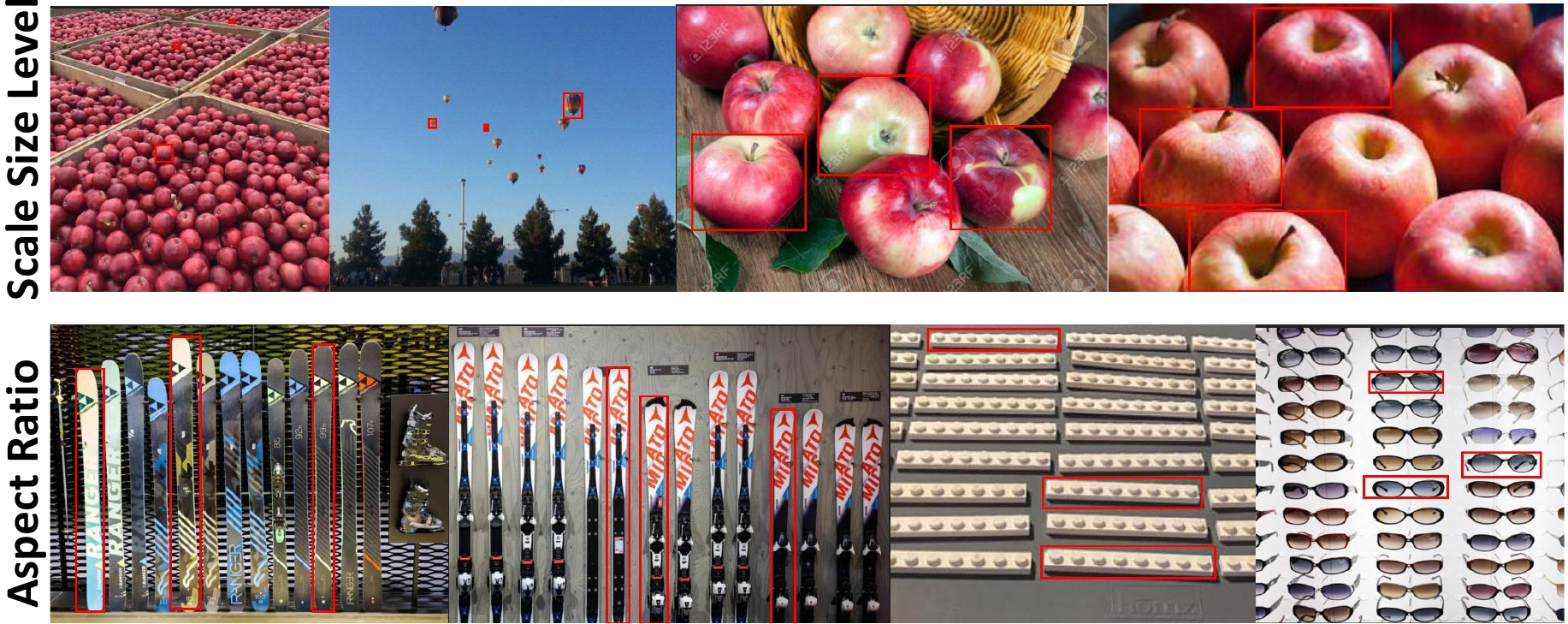}
  \caption{Examples of images with different scale levels and aspect ratio levels. The first and second rows display the variations in scale and in aspect ratio in the FSC147 dataset, respectively.}
  \label{fig:scaleshow}
\end{figure}

\begin{figure}[t]
  \includegraphics[width=\linewidth]{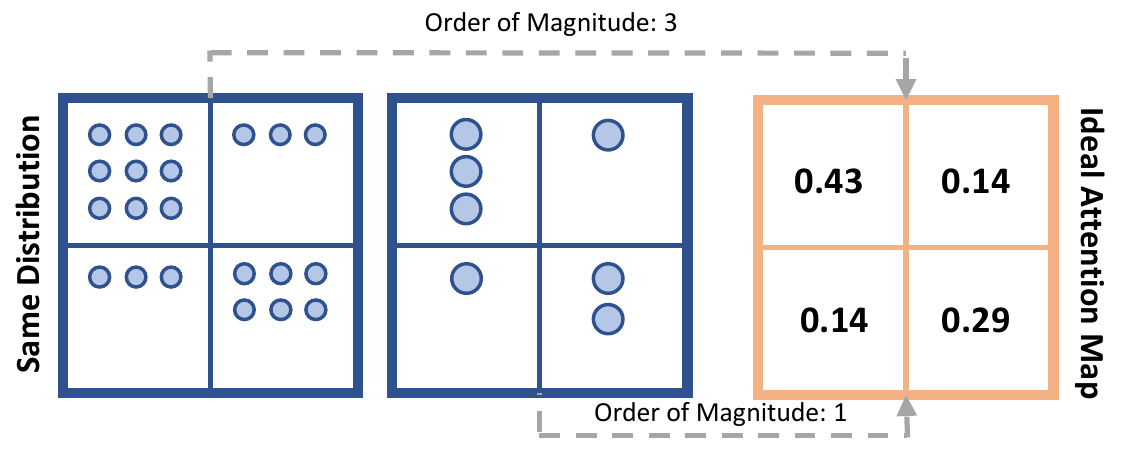}
  \caption{Objective of magnitude embedding. For query images with the same distribution but different counting values, the ideal attention maps should be the same due to normalization, which results in the information loss of the order of magnitude.}
  \label{fig:magnitude}
\end{figure}

While the self-attention mechanism in ViT suits the CAC task, certain restrictions or functions within this structure can result in information loss. First, exemplars must be resized to fit the fixed-size tokens. As shown in Figure~\ref{fig:scaleshow}, the scale size level and aspect ratio of objects vary largely. So the resized exemplars will lose the scale information and complicate the matching process. 
Second, the $\tt softmax$ used to normalize the attention map will weaken the ability to express the number of objects. Per Figure~\ref{fig:magnitude}, for query images with the same distribution but different counting values, the attention maps would be similar due to normalization, which results in the information loss of the order of magnitude.

In order to compensate for the loss and make our CACViT more suitable for the CAC task, we introduce aspect-ratio-aware scale embedding and order-of-magnitude embedding.

\textbf{Aspect-Ratio-Aware Scale Embedding.} While scale-level information has been studied in previous work~\cite{bmnet}, aspect ratio information is often overlooked. Here we propose the scale embedding with further consideration of the aspect ratio. 

\begin{figure}[t]
  \includegraphics[width=1\linewidth]{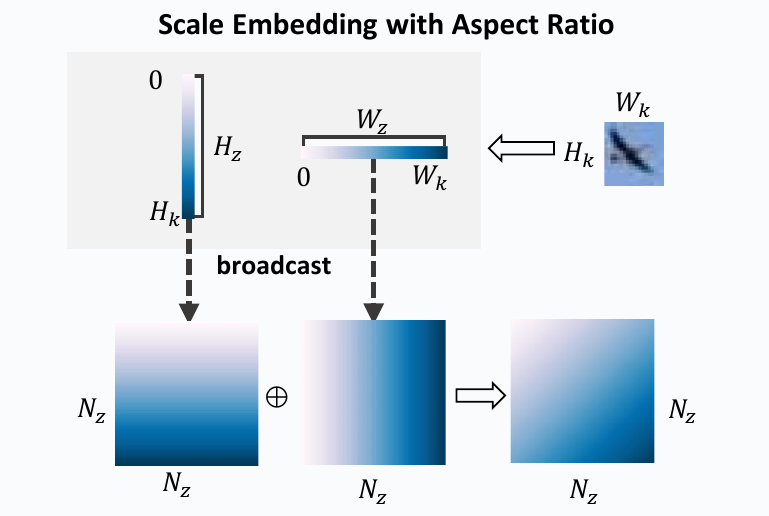}
  \caption{An illustration on how to compute the aspect-ratio-aware scale embedding for exemplar. For a exemplar with width $W_k$ and height $H_k$, which will be resized to fixed width $W_z$ and fixed height $H_z$.} 
  \label{fig:scaleembedding}
\end{figure}

The embedding procedure is shown in Figure~\ref{fig:scaleembedding}. Specifically, given an exemplar $z_k$, whose original width and height are $W_k$ and $H_k$ respectively. After resizing the function, we obtain a fixed size exemplar $z_k \in \mathbb{R}^{W_z \times H_z \times C}$, where $W_z$ and $H_z$ are the fixed-size height and width for every exemplar. First, to keep the width information of the exemplar, we create a linear vector with $W_k$ points between $0$ and $W_k$. Then we can broadcast the vector to acquire a width map $\hat{W}_{k}\in \mathbb{R}^{W_z \times H_z }$. In the same way, we can obtain the height map $\hat{H}_{k}\in \mathbb{R}^{W_z \times H_z }$. Finally, we sum $\hat{W}_{k}$ and $\hat{H}_{k}$ as the scale embedding $S_{k}=\hat{W}_{k} + \hat{H}_{k}$. Then we can concatenate $S_{k}$ with resized exemplar $z_k$ to restore the scale information. Note that, in this way, we also encode the positional information for every patch by scale embedding.

\textbf{Magnitude Embedding.} 
The order-of-magnitude information can be roughly represented by the ratio of the image size to the exemplar size. Specifically, we assume an exemplar is of width $W_k$ and height $H_k$. Considering the patch size is of $W_p \times H_p$ pixels, the maximum capacity of a patch conditioned on the exemplar $z_k$ can be represented by: 
\begin{equation}
M\!E_k = \frac{W_p \times H_p}{W_k\times H_k}\,.
\end{equation}

If we have $K$ exemplars, we can compute the mean magnitude embedding $M\!E$ from exemplars. Then we multiply the embedding with the similarity scores from the attention map to obtain the final similarity map.

\section{Experiments}

\begin{table*}
  \centering
  \renewcommand{\arraystretch}{1}
  \addtolength{\tabcolsep}{0pt}
  \begin{tabular}{@{}lcccccccc@{}}
    \toprule
    Model   & Backbone & Resolution   & Shots & Val MAE & Val RMSE & Test MAE & Test RMSE\\
    \midrule
    FamNet+~\cite{famnet}  & ResNet50 &384   & 1 & 26.55 & 77.01 & 26.76 & 110.95 \\  
    BMNet+~\cite{bmnet}  & ResNet50  & [384,1584]& 1& 17.89 & 61.12 & 16.89 & 96.65\\
    CounTR~\cite{countr}  & ViT-B \& CNN & 384  & 1 & 13.15 & 49.72 & 12.06 & 90.01\\
    CACViT (Ours)  & ViT-B&384& 1 &  \textbf{11.41} & \textbf{41.04} &\textbf{8.62}& \textbf{29.92}\\
    \midrule
    FamNet+~\cite{famnet}  & ResNet-50&384 & 3 & 23.75 & 69.07 & 22.08 & 99.54 \\
    RCAC~\cite{rcac}  & ResNet-50&384  & 3& 20.54 & 60.78 & 20.21 & 81.86\\
    BMNet+~\cite{bmnet}  & ResNet-50 &[384,1584] & 3& 15.74 & 58.53 & 14.62 & 91.83\\
    SAFECount~\cite{safecount}   &ResNet-18 & 512 &3 & 15.28 & 47.20 & 14.32 & 85.54\\
    SPDCN~\cite{spdcn}  & ResNet-18  &576 & 3 & 14.59 & 49.97 & 13.51 & 96.80\\
    CounTR~\cite{countr}  & ViT-B \& CNN  &384& 3 & 13.13 & 49.83 & 11.95 & 91.23\\
    CACViT (Ours)  & ViT-B&384& 3 & \textbf{10.63} & \textbf{37.95} & \textbf{9.13} & \textbf{48.96}\\
    \bottomrule
  \end{tabular}
  \caption{Comparison with the state-of-the-art CAC approaches on the FSC-147 dataset. The upper part of the table presents the results on the 1-shot setting while the lower presents the 3-shots results. CounTR~\cite{countr} need two-stage training regime~\cite{mae}. Note that our 1-shot CACViT outperforms all of the previous methods even in the 3-shots setting.}
  \label{tab:sota}
\end{table*}

\subsection{Implementation Details}
To demonstrate the superiority of our method, we conduct extensive experiments on FSC147~\cite{famnet}, which is the first large-scale dataset for CAC. Similar to~\cite{bmnet}, we also adopt Mean Absolute Error (MAE) and Mean Squared Error (MSE) as the evaluation metrics.

\textbf{Network Structure:} The network takes the image of size $384\times 384$ as the input, which is first split into patches of size $16 \times 16$. Each exemplar is of size $64\times 64$, then split into patches of size $16 \times 16$. Our feature extractor, pre-trained with MAE~\cite{mae}, consists of 12 transformer encoder blocks with a hidden dimension of $768$, and each multi-head self-attention layer contains $12$ heads. The following $3$ extra transformer blocks with a hidden dimension of $512$ are adopted to enhance the feature and reduce the dimension for upsampling. Our regression decoder consists of 4 up-sampling layers with a hidden dimension of $256$ as in CounTR~\cite{countr}. 

\textbf{Training Details:} For fair comparison, we use the same data augmentation, test-time cropping and normalization as CounTR~\cite{countr}. We apply AdamW~\cite{loshchilov2017decoupled} as the optimizer with a batch size of $8$. The model is trained for $200$ epochs with a learning rate of $1e-4$, a weight decay rate of $0.05$, and $10$ epochs for warm up. Our model is trained and tested on NVIDIA GeForce RTX $3090$. Note that the CACViT consumes about $14$GB of memory on a single GPU for $12$ hours to train.

\subsection{Comparison with State of the Art}

\textbf{Quantitative Results.} In this section, we compare our CACViT with state-of-the-art methods on the FSC-147 dataset. As shown in Table~\ref{tab:sota}, CACViT outperforms all the compared methods on both 1-shot and 3-shot settings significantly. In the 1-shot setting, CACViT achieves a relative improvement of $13.23\%$ w.r.t validation MAE and $28.52\%$ w.r.t test MAE compared with CounTR. Besides, CACViT reduces the validation RMSE by $17.45\%$ and the test RMSE by $66.76\%$. In the 3-shot setting, compared with CounTR, CACViT generates a relative improvement of $19.04\%$ w.r.t validation MAE and $23.60\%$ w.r.t test MAE. One can also observe that CACViT reduces the validation RMSE by $23.84\%$ and the test RMSE by $46.33\%$ compared with CounTR, which validates the robustness of our method. 

It is noticed that the performance of the 1-shot is even better than that of the 3-shot in the test set, which is against our intuition. The phenomenon lies in the quality of annotation for dense environments, we provide more detail on the supplementary.

\textbf{Performance of Different Densities.} In order to further validate the performance of the model under different object densities on the 3-shot setting. we divide the test set into two sub-test sets of low and high density, each covers object counts from 8-37 and 37-3701 respectively, and each subset contains roughly 600 images.
As shown in Table~\ref{tab:differentdensity}, our method outperforms other methods in both the low-density and high-density environments, which validates the robustness of our method.

\begin{table}[t]
  \centering
  \renewcommand{\arraystretch}{1.1}
  \addtolength{\tabcolsep}{-4pt}
  \begin{tabular}{ccccccc}
    \toprule
    method & low MAE  & low RMSE & high MAE  & high RMSE  \\
    \midrule
    BMNet+ &  6.23 & 29.96  & 23.21 &   127.62\\
    SPDCN & 5.06 &   31.53  &  22.25 & 133.59\\
    CounTR &  5.44 &  36.96    & 18.56 &  124.11  \\
    CACViT  & \textbf{ 3.14} &  \textbf{6.26}  & \textbf{15.33} & \textbf{69.54}\\
  \bottomrule
\end{tabular}
\caption{Comparison with the state-of-the-art CAC approaches on two sub-test sets of low and high density, each covers object counts from 8-37 and 37-3701 respectively, and each subset contains roughly 600 images. }
  \label{tab:differentdensity}
\end{table}

\textbf{Running Cost Evaluation.} 
To verify the efficiency of our method, we compare our model size, floating point operations (FLOPS) with CounTR~\cite{countr}, which employs the same backbone as CACViT. The results are reported in Table~\ref{tab:cost}. Note that CounTR does not report its epochs for finetuning in their paper, so we get the data from its official GitHub page. Our training setting is similar to CounTR. However, 1) we exclude the self-supervised~\cite{mae} pre-training on the FSC147 dataset compared with CounTR, and 2) we do not require an additional CNN to extract features of exemplars. One can easily observe that our CACViT can significantly reduce the training epochs from $1300$ to $200$ and simplify the training strategy into one stage while keeping the model size and FLOPs comparable to that of CounTR.

\begin{figure}[t]
  \centering
  \includegraphics[width=\linewidth]{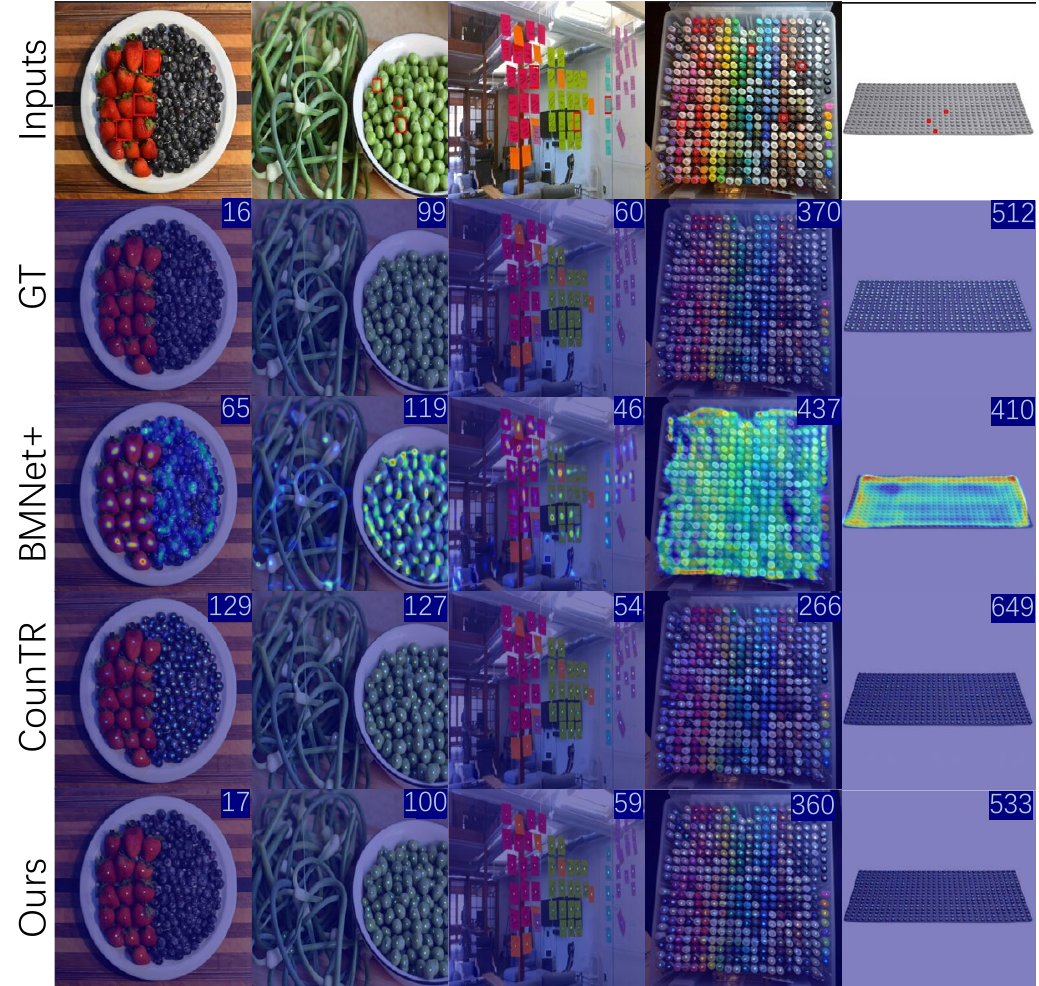}
  \caption{Qualitative results on the FSC147 dataset. Different challenges are shown in the selected inputs, including images with multiple classes, dense environments, and colorful environments. Our method consistently outperforms previous methods with more precise locations.}
  \label{fig:qualitative}
\end{figure}

\textbf{Qualitative Analysis.} For qualitative analysis, we compare our CACViT with the state-of-the-art CNN-based model BMNet+ and state-of-the-art ViT-based model CounTR. As shown in Fig~\ref{fig:qualitative}, CACViT outperforms other methods in different scenes, including images with multiple classes, or in dense environments and colorful environments. For hard cases involving interference from other classes (the 1st and 2nd column), our method allows better learning of the characteristics of the objects to be counted. 

\begin{table}
  \centering
  \renewcommand{\arraystretch}{1.1}
  \addtolength{\tabcolsep}{-2pt}
  \begin{tabular}{cccc}
    \toprule
    method & Model Size (M) & GFLOPs & \#epochs \\
    \midrule
    CounTR pretrain &  112M &  27G & 300\\
    CounTR finetune &  99M &  84G & 1000*\\
    CACViT (Ours) &  99M & 89G & 200\\
  \bottomrule
\end{tabular}
\caption{Comparison of the model size and FLOPs. * indicates that we obtain the \#epochs on its official project page, which is not mentioned on paper.}
  \label{tab:cost}
\end{table}

\subsection{Ablation Study}

We perform the ablation study on FSC147 and provide quantitative results in Table~\ref{tab:ablation}. The plain version of CACViT is employed as the baseline which only consists of standard feature extraction and matching blocks by self-attention, and the $A_{class}$ part in attention map is masked (B1 and B6).

\textbf{Additional class information of $A_{class}$.} By comparing B6 and B7 in Table~\ref{tab:ablation}, one can observe that background suppression can introduce a relative improvement of $15.44\%$ w.r.t. MAE and $23.12\%$ w.r.t. RMSE on the test set. The performance boost indicates that exemplars can provide valid information for query features.

\textbf{Aspect-ratio-aware scale embedding.} As shown in Table~\ref{tab:ablation}, the aspect-ratio-aware scale embedding brings a relative improvement of $14.44\%$ MAE and $16.87\%$ RMSE on the validation set and $5.80\%$ MAE and $13.69\%$ RMSE on the test set (cf. B2 \textit{vs.} B3 and B7 \textit{vs.} B8), which verifies that the aspect-ratio-aware scale embedding can restore original size information for exemplars, and therefore matters in CAC task. Note that, the performance on RMSE further demonstrates that the SE module can boost the robustness of the model. Besides, we provide a comparison with other scale embedding methods in the supplementary material.

\begin{table}[t]
  \centering
  \renewcommand{\arraystretch}{1}
  \addtolength{\tabcolsep}{2pt}
  \begin{tabular}{@{}lccccc@{}}
    \toprule
    No. & CLS &SE & ME &Val MAE &Val RMSE\\
    \midrule
    B1 &  & &  & 13.00&48.37\\
    B2 &  \checkmark &  &&12.40 & 48.62\\
    B3 &  \checkmark &   \checkmark&&10.61&40.42\\
    B4 &  \checkmark&   &\checkmark&12.32&42.42\\
    B5 &  \checkmark &   \checkmark&   \checkmark&\textbf{10.62}&\textbf{37.95}\\
    \midrule
    No. & CLS &SE &ME &Test MAE &Test RMSE\\
    \midrule
    B6 &  & &  & 13.67&117.60\\
    B7 &  \checkmark &  & &11.56 & 90.41\\
    B8 &  \checkmark &  \checkmark& &10.89&78.03\\
    B9 & \checkmark &  & \checkmark& 11.61&83.06\\
    B10 &  \checkmark &  \checkmark&  \checkmark&\textbf{9.13}&\textbf{48.96}\\
  \bottomrule
\end{tabular}
\caption{Ablation study on the FSC147 dataset. CLS denotes the additional component $A_{class}$ in self-attention compared with previous methods, SE refers to scale embedding, and ME denotes magnitude embedding. 
}
  \label{tab:ablation}
\end{table}

\textbf{Magnitude Embedding.} The comparison of B2 \textit{vs.} B4 and B7 \textit{vs.} B9 demonstrates that magnitude embedding improves the validation RMSE by 6.2 and test RMSE by 7.35. However, the performance boost on MAE is marginal or even declines slightly. The reason may lies in that resized exemplars provide a wrong prior of object size which confuses the model. 

\textbf{Cooperation between Scale Embedding and Magnitude Embedding.} By Comparing B8 with B10 in Table~\ref{tab:ablation}, one can observe that magnitude embedding can lead to a relative improvement of $21.36\%$ w.r.t. MAE and $41.05\%$ w.r.t. RMSE on the test set. Note that the improvement is much more significant compared with the situation without scale embedding, which indicates that the combination of scale embedding and magnitude embedding can further boost the performance. The synergy between the scale embedding and magnitude embedding can provide scale prior for input and output separately, therefore CACViT can capture the scale prior all the time.

\subsection{Cross-Dataset Generalization}

\begin{table}[t]
  \centering
  \renewcommand{\arraystretch}{1}
  \addtolength{\tabcolsep}{7pt}
  \begin{tabular}{@{}lccc@{}}
    \toprule
    Method & Fine-tuned &MAE &RMSE\\
    \midrule
    FamNet+ &  × & 28.84&44.47\\
    RCAC~ &  × &17.98 & 24.21\\
    SAFECount &  ×&16.66&24.08\\
    BMNet+ &  ×&10.44&13.77\\
    CACViT (Ours) &  × &\textbf{8.30}&\textbf{11.18}\\
    \midrule
    FamNet+ &  \checkmark & 18.19 &33.66\\
    RCAC &  \checkmark &13.62 & 19.08\\
    SAFECount &  \checkmark&5.33&7.04\\
    BMNet+ &  \checkmark&5.76&7.83\\
    CounTR &  \checkmark  &5.75&7.45\\
    CACViT (Ours) &  \checkmark &\textbf{4.91}&\textbf{6.49}\\
  \bottomrule
\end{tabular}
\vspace{4pt}

\caption{Generalization performance on the CARPK dataset. All models are pre-trained on the FSC147 dataset without the 'cars' class. "fine-tuned" denotes whether the pre-trained models are further fine-tuned on the CARPK dataset.}
  \label{tab:carpk}
\end{table}

We test the generality of the model on a car counting dataset CARPK\cite{carpk}. 
We exclude the “cars” category within FSC147 to ensure that training and test categories have no overlap. Results are reported in Table~\ref{tab:carpk}. 
One can observe that our model exhibits strong generality. Compared with BMNet+\cite{bmnet}, CACViT generates a relative performance gain of $20.50\%$ MAE.

\section{Conclusions}

In this work, we propose a simple yet efficient ViT-based model CACViT for CAC. Specifically, we show that the ViT is naturally suitable for the CAC task from a decoupled view. And we propose a ViT-based extract-and-match paradigm for CAC. Then we introduce aspect-ratio-aware scale embedding and magnitude embedding to compensate for the information loss. Our CACViT achieves stat-of-the-art results on FSC147, and we also verify the generality on CARPK.

\bibliography{aaai24}

\begin{thebibliography}{36}
\providecommand{\natexlab}[1]{#1}

\bibitem[{Abousamra et~al.(2021)Abousamra, Hoai, Samaras, and
  Chen}]{abousamra2021localization}
Abousamra, S.; Hoai, M.; Samaras, D.; and Chen, C. 2021.
\newblock Localization in the crowd with topological constraints.
\newblock In \emph{Proceedings of the AAAI Conference on Artificial
  Intelligence}, volume~35, 872--881.

\bibitem[{Cao et~al.(2018)Cao, Wang, Zhao, and Su}]{cao2018scale}
Cao, X.; Wang, Z.; Zhao, Y.; and Su, F. 2018.
\newblock Scale aggregation network for accurate and efficient crowd counting.
\newblock In \emph{Proceedings of the European conference on computer vision
  (ECCV)}, 734--750.

\bibitem[{Chen et~al.(2022)Chen, Wu, Wang, Hu, Hu, Ding, Cheng, and
  Wang}]{mixformer}
Chen, Q.; Wu, Q.; Wang, J.; Hu, Q.; Hu, T.; Ding, E.; Cheng, J.; and Wang, J.
  2022.
\newblock MixFormer: Mixing Features Across Windows and Dimensions.
\newblock In \emph{Proceedings of the IEEE/CVF Conference on Computer Vision
  and Pattern Recognition (CVPR)}, 5249--5259.

\bibitem[{Cheng et~al.(2022)Cheng, Dai, Li, Song, Wu, and
  Hauptmann}]{cheng2022rethinking}
Cheng, Z.-Q.; Dai, Q.; Li, H.; Song, J.; Wu, X.; and Hauptmann, A.~G. 2022.
\newblock Rethinking spatial invariance of convolutional networks for object
  counting.
\newblock In \emph{Proceedings of the IEEE/CVF Conference on Computer Vision
  and Pattern Recognition}, 19638--19648.

\bibitem[{Gong et~al.(2022)Gong, Zhang, Yang, Dai, and Schiele}]{rcac}
Gong, S.; Zhang, S.; Yang, J.; Dai, D.; and Schiele, B. 2022.
\newblock Class-Agnostic Object Counting Robust to Intraclass Diversity.
\newblock In \emph{Computer Vision--ECCV 2022: 17th European Conference, Tel
  Aviv, Israel, October 23--27, 2022, Proceedings, Part XXXIII}, 388--403.
  Springer.

\bibitem[{Han et~al.(2022)Han, Ma, Huang, Chen, and Chang}]{crossdet}
Han, G.; Ma, J.; Huang, S.; Chen, L.; and Chang, S.-F. 2022.
\newblock Few-Shot Object Detection With Fully Cross-Transformer.
\newblock In \emph{Proceedings of the IEEE/CVF Conference on Computer Vision
  and Pattern Recognition (CVPR)}, 5321--5330.

\bibitem[{He et~al.(2022)He, Chen, Xie, Li, Doll{\'a}r, and Girshick}]{mae}
He, K.; Chen, X.; Xie, S.; Li, Y.; Doll{\'a}r, P.; and Girshick, R. 2022.
\newblock Masked autoencoders are scalable vision learners.
\newblock In \emph{Proceedings of the IEEE/CVF Conference on Computer Vision
  and Pattern Recognition}, 16000--16009.

\bibitem[{He et~al.(2016)He, Zhang, Ren, and Sun}]{he2016deep}
He, K.; Zhang, X.; Ren, S.; and Sun, J. 2016.
\newblock Deep residual learning for image recognition.
\newblock In \emph{Proceedings of the IEEE conference on computer vision and
  pattern recognition}, 770--778.

\bibitem[{He et~al.(2021)He, Ma, Wei, Hong, Ke, and Gong}]{he2021error}
He, Y.; Ma, Z.; Wei, X.; Hong, X.; Ke, W.; and Gong, Y. 2021.
\newblock Error-aware density isomorphism reconstruction for unsupervised
  cross-domain crowd counting.
\newblock In \emph{Proceedings of the AAAI conference on artificial
  intelligence}, volume~35, 1540--1548.

\bibitem[{Hsieh, Lin, and Hsu(2017)}]{carpk}
Hsieh, M.-R.; Lin, Y.-L.; and Hsu, W.~H. 2017.
\newblock Drone-based object counting by spatially regularized regional
  proposal network.
\newblock In \emph{Proceedings of the IEEE international conference on computer
  vision}, 4145--4153.

\bibitem[{Idrees et~al.(2018)Idrees, Tayyab, Athrey, Zhang, Al-Maadeed,
  Rajpoot, and Shah}]{idrees2018composition}
Idrees, H.; Tayyab, M.; Athrey, K.; Zhang, D.; Al-Maadeed, S.; Rajpoot, N.; and
  Shah, M. 2018.
\newblock Composition loss for counting, density map estimation and
  localization in dense crowds.
\newblock In \emph{Proceedings of the European conference on computer vision
  (ECCV)}, 532--546.

\bibitem[{Kim, Son, and Kim(2021)}]{ViLT}
Kim, W.; Son, B.; and Kim, I. 2021.
\newblock ViLT: Vision-and-Language Transformer Without Convolution or Region
  Supervision.
\newblock In Meila, M.; and Zhang, T., eds., \emph{Proceedings of the 38th
  International Conference on Machine Learning}, volume 139 of
  \emph{Proceedings of Machine Learning Research}, 5583--5594. PMLR.

\bibitem[{Laradji et~al.(2018)Laradji, Rostamzadeh, Pinheiro, Vazquez, and
  Schmidt}]{laradji2018blobs}
Laradji, I.~H.; Rostamzadeh, N.; Pinheiro, P.~O.; Vazquez, D.; and Schmidt, M.
  2018.
\newblock Where are the blobs: Counting by localization with point supervision.
\newblock In \emph{Proceedings of the european conference on computer vision
  (ECCV)}, 547--562.

\bibitem[{Li et~al.(2022)Li, Mao, Girshick, and He}]{vitdet}
Li, Y.; Mao, H.; Girshick, R.; and He, K. 2022.
\newblock Exploring Plain Vision Transformer Backbones for Object Detection.
\newblock arXiv:2203.16527.

\bibitem[{Lin et~al.(2022)Lin, Yang, Ma, Gao, Liu, Liu, Hou, Yi, and
  Chan}]{spdcn}
Lin, W.; Yang, K.; Ma, X.; Gao, J.; Liu, L.; Liu, S.; Hou, J.; Yi, S.; and
  Chan, A.~B. 2022.
\newblock Scale-Prior Deformable Convolution for Exemplar-Guided Class-Agnostic
  Counting.

\bibitem[{Lin et~al.(2023)Lin, Yuan, Zhang, Li, Zheng, and Hu}]{detrvit}
Lin, Y.; Yuan, Y.; Zhang, Z.; Li, C.; Zheng, N.; and Hu, H. 2023.
\newblock DETR Doesn't Need Multi-Scale or Locality Design.
\newblock arXiv:2308.01904.

\bibitem[{Liu et~al.(2022)Liu, Zhong, Zisserman, and Xie}]{countr}
Liu, C.; Zhong, Y.; Zisserman, A.; and Xie, W. 2022.
\newblock Countr: Transformer-based generalised visual counting.
\newblock \emph{arXiv preprint arXiv:2208.13721}.

\bibitem[{Loshchilov and Hutter(2017)}]{loshchilov2017decoupled}
Loshchilov, I.; and Hutter, F. 2017.
\newblock Decoupled weight decay regularization.
\newblock \emph{arXiv preprint arXiv:1711.05101}.

\bibitem[{Lu, Xie, and Zisserman(2019)}]{lu2019class}
Lu, E.; Xie, W.; and Zisserman, A. 2019.
\newblock Class-agnostic counting.
\newblock In \emph{Computer Vision--ACCV 2018: 14th Asian Conference on
  Computer Vision, Perth, Australia, December 2--6, 2018, Revised Selected
  Papers, Part III 14}, 669--684. Springer.

\bibitem[{Lu et~al.(2017)Lu, Cao, Xiao, Zhuang, and Shen}]{lu2017tasselnet}
Lu, H.; Cao, Z.; Xiao, Y.; Zhuang, B.; and Shen, C. 2017.
\newblock TasselNet: counting maize tassels in the wild via local counts
  regression network.
\newblock \emph{Plant methods}, 13(1): 1--17.

\bibitem[{Madec et~al.(2019)Madec, Jin, Lu, De~Solan, Liu, Duyme, Heritier, and
  Baret}]{madec2019ear}
Madec, S.; Jin, X.; Lu, H.; De~Solan, B.; Liu, S.; Duyme, F.; Heritier, E.; and
  Baret, F. 2019.
\newblock Ear density estimation from high resolution RGB imagery using deep
  learning technique.
\newblock \emph{Agricultural and forest meteorology}, 264: 225--234.

\bibitem[{Onoro-Rubio and L{\'o}pez-Sastre(2016)}]{onoro2016towards}
Onoro-Rubio, D.; and L{\'o}pez-Sastre, R.~J. 2016.
\newblock Towards perspective-free object counting with deep learning.
\newblock In \emph{Computer Vision--ECCV 2016: 14th European Conference,
  Amsterdam, The Netherlands, October 11--14, 2016, Proceedings, Part VII 14},
  615--629. Springer.

\bibitem[{Ranjan et~al.(2021)Ranjan, Sharma, Nguyen, and Hoai}]{famnet}
Ranjan, V.; Sharma, U.; Nguyen, T.; and Hoai, M. 2021.
\newblock Learning to count everything.
\newblock In \emph{Proceedings of the IEEE/CVF Conference on Computer Vision
  and Pattern Recognition}, 3394--3403.

\bibitem[{Shi et~al.(2022)Shi, Lu, Feng, Liu, and Cao}]{bmnet}
Shi, M.; Lu, H.; Feng, C.; Liu, C.; and Cao, Z. 2022.
\newblock Represent, compare, and learn: A similarity-aware framework for
  class-agnostic counting.
\newblock In \emph{Proceedings of the IEEE/CVF Conference on Computer Vision
  and Pattern Recognition}, 9529--9538.

\bibitem[{Shu et~al.(2022)Shu, Wan, Tan, Kwong, and Chan}]{shu2022crowd}
Shu, W.; Wan, J.; Tan, K.~C.; Kwong, S.; and Chan, A.~B. 2022.
\newblock Crowd counting in the frequency domain.
\newblock In \emph{Proceedings of the IEEE/CVF Conference on Computer Vision
  and Pattern Recognition}, 19618--19627.

\bibitem[{Touvron et~al.(2023)Touvron, Martin, Stone, Albert, Almahairi,
  Babaei, Bashlykov, Batra, Bhargava, Bhosale, Bikel, Blecher, Ferrer, Chen,
  Cucurull, Esiobu, Fernandes, Fu, Fu, Fuller, Gao, Goswami, Goyal, Hartshorn,
  Hosseini, Hou, Inan, Kardas, Kerkez, Khabsa, Kloumann, Korenev, Koura,
  Lachaux, Lavril, Lee, Liskovich, Lu, Mao, Martinet, Mihaylov, Mishra,
  Molybog, Nie, Poulton, Reizenstein, Rungta, Saladi, Schelten, Silva, Smith,
  Subramanian, Tan, Tang, Taylor, Williams, Kuan, Xu, Yan, Zarov, Zhang, Fan,
  Kambadur, Narang, Rodriguez, Stojnic, Edunov, and Scialom}]{touvron2023llama}
Touvron, H.; Martin, L.; Stone, K.; Albert, P.; Almahairi, A.; Babaei, Y.;
  Bashlykov, N.; Batra, S.; Bhargava, P.; Bhosale, S.; Bikel, D.; Blecher, L.;
  Ferrer, C.~C.; Chen, M.; Cucurull, G.; Esiobu, D.; Fernandes, J.; Fu, J.; Fu,
  W.; Fuller, B.; Gao, C.; Goswami, V.; Goyal, N.; Hartshorn, A.; Hosseini, S.;
  Hou, R.; Inan, H.; Kardas, M.; Kerkez, V.; Khabsa, M.; Kloumann, I.; Korenev,
  A.; Koura, P.~S.; Lachaux, M.-A.; Lavril, T.; Lee, J.; Liskovich, D.; Lu, Y.;
  Mao, Y.; Martinet, X.; Mihaylov, T.; Mishra, P.; Molybog, I.; Nie, Y.;
  Poulton, A.; Reizenstein, J.; Rungta, R.; Saladi, K.; Schelten, A.; Silva,
  R.; Smith, E.~M.; Subramanian, R.; Tan, X.~E.; Tang, B.; Taylor, R.;
  Williams, A.; Kuan, J.~X.; Xu, P.; Yan, Z.; Zarov, I.; Zhang, Y.; Fan, A.;
  Kambadur, M.; Narang, S.; Rodriguez, A.; Stojnic, R.; Edunov, S.; and
  Scialom, T. 2023.
\newblock Llama 2: Open Foundation and Fine-Tuned Chat Models.
\newblock arXiv:2307.09288.

\bibitem[{Vaswani et~al.(2017)Vaswani, Shazeer, Parmar, Uszkoreit, Jones,
  Gomez, Kaiser, and Polosukhin}]{vaswani2017attention}
Vaswani, A.; Shazeer, N.; Parmar, N.; Uszkoreit, J.; Jones, L.; Gomez, A.~N.;
  Kaiser, {\L}.; and Polosukhin, I. 2017.
\newblock Attention is all you need.
\newblock \emph{Advances in neural information processing systems}, 30.

\bibitem[{Wang et~al.(2020)Wang, Gao, Lin, and Li}]{wang2020nwpu}
Wang, Q.; Gao, J.; Lin, W.; and Li, X. 2020.
\newblock NWPU-crowd: A large-scale benchmark for crowd counting and
  localization.
\newblock \emph{IEEE transactions on pattern analysis and machine
  intelligence}, 43(6): 2141--2149.

\bibitem[{Xu et~al.(2022)Xu, Zhang, ZHANG, and Tao}]{vitpose}
Xu, Y.; Zhang, J.; ZHANG, Q.; and Tao, D. 2022.
\newblock ViTPose: Simple Vision Transformer Baselines for Human Pose
  Estimation.
\newblock In Koyejo, S.; Mohamed, S.; Agarwal, A.; Belgrave, D.; Cho, K.; and
  Oh, A., eds., \emph{Advances in Neural Information Processing Systems},
  volume~35, 38571--38584. Curran Associates, Inc.

\bibitem[{Xu et~al.(2023)Xu, Zhang, Zhang, and Tao}]{xu2023vitpose}
Xu, Y.; Zhang, J.; Zhang, Q.; and Tao, D. 2023.
\newblock ViTPose++: Vision Transformer Foundation Model for Generic Body Pose
  Estimation.
\newblock arXiv:2212.04246.

\bibitem[{Yang et~al.(2021)Yang, Su, Hsu, and Chen}]{yang2021class}
Yang, S.-D.; Su, H.-T.; Hsu, W.~H.; and Chen, W.-C. 2021.
\newblock Class-agnostic few-shot object counting.
\newblock In \emph{Proceedings of the IEEE/CVF Winter Conference on
  Applications of Computer Vision}, 870--878.

\bibitem[{Yao et~al.(2023)Yao, Wang, Yang, and Wang}]{ViTMatte}
Yao, J.; Wang, X.; Yang, S.; and Wang, B. 2023.
\newblock ViTMatte: Boosting Image Matting with Pretrained Plain Vision
  Transformers.
\newblock arXiv:2305.15272.

\bibitem[{You et~al.(2023)You, Yang, Luo, Lu, Cui, and Le}]{safecount}
You, Z.; Yang, K.; Luo, W.; Lu, X.; Cui, L.; and Le, X. 2023.
\newblock Few-shot object counting with similarity-aware feature enhancement.
\newblock In \emph{Proceedings of the IEEE/CVF Winter Conference on
  Applications of Computer Vision}, 6315--6324.

\bibitem[{Yu et~al.(2022)Yu, Luo, Zhou, Si, Zhou, Wang, Feng, and
  Yan}]{yu2022metaformer}
Yu, W.; Luo, M.; Zhou, P.; Si, C.; Zhou, Y.; Wang, X.; Feng, J.; and Yan, S.
  2022.
\newblock MetaFormer Is Actually What You Need for Vision.
\newblock arXiv:2111.11418.

\bibitem[{Zhang et~al.(2015)Zhang, Li, Wang, and Yang}]{zhang2015cross}
Zhang, C.; Li, H.; Wang, X.; and Yang, X. 2015.
\newblock Cross-scene crowd counting via deep convolutional neural networks.
\newblock In \emph{Proceedings of the IEEE conference on computer vision and
  pattern recognition}, 833--841.

\bibitem[{Zou et~al.(2021)Zou, Qu, Zhou, Xu, Ye, Wu, and Ye}]{zou2021coarse}
Zou, Z.; Qu, X.; Zhou, P.; Xu, S.; Ye, X.; Wu, W.; and Ye, J. 2021.
\newblock Coarse to fine: Domain adaptive crowd counting via adversarial
  scoring network.
\newblock In \emph{Proceedings of the 29th ACM International Conference on
  Multimedia}, 2185--2194.

\end{thebibliography}

\end{document}